%% file: paper.tex
\documentclass{article}
\usepackage{ijcai24}

\usepackage{times}
\usepackage{soul}
\usepackage{url}
\usepackage[hidelinks]{hyperref}
\usepackage[utf8]{inputenc}
\usepackage[small]{caption}
\usepackage{graphicx}
\graphicspath{{./images/}} %
\usepackage[svgnames,table,rgb,dvipsnames]{xcolor} %
\usepackage{amsmath}
\usepackage{amsthm}
\usepackage{amssymb}
\usepackage{booktabs}
\usepackage{algorithm}
\usepackage{algorithmicx}
\usepackage[noend]{algpseudocode}
\usepackage[switch]{lineno}
\usepackage{comment}
\usepackage{nicefrac}
\usepackage{amsbsy}

\makeatletter
\newcommand*{\algrule}[1][\algorithmicindent]{\makebox[#1][l]{\hspace*{.5em}\vrule height .73\baselineskip depth .27\baselineskip}}%

\newcount\ALG@printindent@tempcnta
\def\ALG@printindent{%
    \ifnum \theALG@nested>0%
        \ifx\ALG@text\ALG@x@notext%
            \addvspace{-3pt}%
        \else
            \unskip
            \ALG@printindent@tempcnta=1
            \loop
                \algrule[\csname ALG@ind@\the\ALG@printindent@tempcnta\endcsname]%
                \advance \ALG@printindent@tempcnta 1
            \ifnum \ALG@printindent@tempcnta<\numexpr\theALG@nested+1\relax%
            \repeat
        \fi
    \fi
    }%
\usepackage{etoolbox}
\patchcmd{\ALG@doentity}{\noindent\hskip\ALG@tlm}{\ALG@printindent}{}{\errmessage{failed to patch}}
\makeatother

\usepackage{mathtools}

\DeclareMathOperator*{\argmax}{argmax}
\usepackage{algpseudocode}
\usepackage{todonotes}
\presetkeys{todonotes}{inline}{}
\usepackage{tabulary}
\usepackage{pgfplots}
\pgfplotsset{compat=1.18,
    width=0.25\textwidth,
    height=0.25\textwidth,
}
\usepgfplotslibrary{fillbetween}
\usetikzlibrary{matrix}
\usepgfplotslibrary{groupplots}
\pgfplotsset{
    legend style={
        at={(0.5,-0.24)},
        anchor=north,
        legend columns=-1,
    },
}
\pgfplotsset{every tick label/.append style={font=\tiny}}

\newcommand{\ours}{AmEx-MCTS}
\newcommand{\ourmax}{Am\AE{}x-MCTS}
\newcommand{\placeholder}{\cdot}

\newcommand{\qvalue}{Q-value}

\usepackage[sort&compress,capitalise,noabbrev]{cleveref}
\crefname{line}{line}{lines} %

\title{Amplifying Exploration in Monte-Carlo Tree Search by Focusing on the Unknown}

\author{
Cedric Derstroff$^{1,2,\ast}$\and
Jannis Brugger$^{1,2,}$\thanks{These authors contributed equally.}\and
Jannis Blüml$^{1,2}$\and
Mira Mezini$^{1,2,3}$\and\\
Stefan Kramer$^4$\And
Kristian Kersting$^{1,2,5,6}$
\affiliations
$^1$ Technical University Darmstadt, Germany \\
$^2$ Hessian Center for Artificial Intelligence (hessian.AI), Darmstadt, Germany \\
$^3$ National Research Center for Applied Cybersecurity ATHENE, Darmstadt, Germany \\
$^4$ Johannes Gutenberg-Universität Mainz, Germany\\
$^5$ Centre for Cognitive Science, Darmstadt, Germany \\
$^6$ German Research Center for Artificial Intelligence (DFKI), Darmstadt, Germany  \\
\emails
\{cedric.derstroff, jannis.brugger, jannis.blueml, mira.mezini, kristian.kersting\}@tu-darmstadt.de, kramer@informatik.uni-mainz.de
}

\input{chapters/plots}

\begin{document}

\maketitle

\begin{abstract}
 Monte-Carlo tree search (MCTS) is an effective anytime algorithm with a vast amount of applications. It strategically allocates computational resources to focus on promising segments of the search tree, making it a very attractive search algorithm in large search spaces. However, it often expends its limited resources on reevaluating previously explored regions when they remain the most promising path. Our proposed methodology, denoted as \ours{}, solves this problem by introducing a novel MCTS formulation. Central to \ours{} is the decoupling of value updates, visit count updates, and the selected path during the tree search, thereby enabling the exclusion of already explored subtrees or leaves. This segregation preserves the utility of visit counts for both exploration-exploitation balancing and quality metrics within MCTS. The resultant augmentation facilitates in a considerably broader search using identical computational resources, preserving the essential characteristics of MCTS. The expanded coverage not only yields more precise estimations but also proves instrumental in larger and more complex problems. Our empirical evaluation demonstrates the superior performance of \ours{}, surpassing classical MCTS and related approaches by a substantial margin.
\end{abstract}

\input{chapters/introduction}
\input{chapters/method}

\input{chapters/experiments}

\input{chapters/related_work}

\input{chapters/limitations}
\input{chapters/conclusion}

\section*{Acknowledgements}
This research project was partly funded by the Hessian Ministry of Science and the Arts (HMWK) within the projects ``The Third Wave of Artificial Intelligence - 3AI'' and hessian.AI.

\bibliographystyle{named}
\bibliography{bib}

\clearpage
\appendix
\input{chapters/appendix.tex}

\end{document}

%% file: chapters/plots.tex
\definecolor{mcts}{HTML}{9B3D12}
\definecolor{mctst}{HTML}{F991CC}
\definecolor{mctst+}{HTML}{F2C57C}
\definecolor{mctsendgame}{HTML}{272D2D}
\definecolor{ourmax}{HTML}{623CEA}

\newcommand{\nsims}{$n_{sims}$}

\newcommand{\chain}{
    \begin{tikzpicture}
        \begin{groupplot}[
            group style={group size=4 by 1},
                ymin=-0.1,
                ymax=1.1,
                xmin=-0.5,
                xmax=5.5,
                grid=major,
                extra y ticks={0.25,0.75},
                xtick={0,...,5},
                xticklabels={5, 10, 25, 50, 100, 250},
        ]
            \nextgroupplot[
                title=Chain-10,
                legend to name=zelda,
                ylabel=Return,
                xlabel=\nsims,
            ]{
                \addplot[mcts, mark=*] coordinates {
                    (0, 5/287)
                    (1, 10/287)
                    (2, 38/287)
                    (3, 51/287)
                    (4, 79/287)
                    (5, 147/287)
                };
                \addlegendentry{MCTS};
                \addplot[mctst, mark=*] coordinates {
                    (0, 38/287)
                    (1, 104/287)
                    (2, 182/287)
                    (3, 287/287)
                    (4, 287/287)
                    (5, 287/287)
                };
                \addlegendentry{MCTS-T};
                \addplot[mctsendgame, mark=*] coordinates {
                    (0, 0.08)
                    (1, 0.56)
                    (2, 1)
                    (3, 1)
                    (4, 1)
                    (5, 1)
                };
                \addlegendentry{\ours{} (ours)};
                \addplot[ourmax, mark=*] coordinates {
                    (0, 0.08)
                    (1, 0.6)
                    (2, 1)
                    (3, 1)
                    (4, 1)
                    (5, 1)
                };
                \addlegendentry{\ourmax{} (ours)};
                \coordinate (top) at (rel axis cs:1,0);
            }
            \nextgroupplot[
                title=Chain-25,
            ]{
                \addplot[mcts, mark=*] coordinates {
                    (0, 0/287)
                    (1, 0/287)
                    (2, 0/287)
                    (3, 0/287)
                    (4, 0/287)
                    (5, 0/287)
                };
                \addplot[mctst, mark=*] coordinates {
                    (0, 0.5/287)
                    (1, 14.5/287)
                    (2, 75.5/287)
                    (3, 151/287)
                    (4, 262/287)
                    (5, 287/287)
                };
                \addplot[mctsendgame, mark=*] coordinates {
                    (0, 0)
                    (1, 0.16)
                    (2, 0.4)
                    (3, 1)
                    (4, 1)
                    (5, 1)
                };
                \addplot[ourmax, mark=*] coordinates {
                    (0, 0)
                    (1, 0.16)
                    (2, 0.4)
                    (3, 1)
                    (4, 1)
                    (5, 1)
                };
            }
            \nextgroupplot[
                title=Chain-50,
            ]{
                \addplot[mcts, mark=*] coordinates {
                    (0, 0/287)
                    (1, 0/287)
                    (2, 0/287)
                    (3, 0/287)
                    (4, 0/287)
                    (5, 0/287)
                };
                \addplot[mctst, mark=*] coordinates {
                    (0, 0/287)
                    (1, 0/287)
                    (2, 21/287)
                    (3, 74/287)
                    (4, 145.5/287)
                    (5, 287/287)
                };
                \addplot[mctsendgame, mark=*] coordinates {
                    (0, 0)
                    (1, 0)
                    (2, 0.08)
                    (3, 0.52)
                    (4, 1)
                    (5, 1)
                };
                \addplot[ourmax, mark=*] coordinates {
                    (0, 0)
                    (1, 0)
                    (2, 0.08)
                    (3, 0.52)
                    (4, 1)
                    (5, 1)
                };
            }
            \nextgroupplot[
                title=Chain-100,
            ]{
                \addplot[mcts, mark=*] coordinates {
                    (0, 0/287)
                    (1, 0/287)
                    (2, 0/287)
                    (3, 0/287)
                    (4, 0/287)
                    (5, 0/287)
                };
                \addplot[mctst, mark=*] coordinates {
                    (0, 0/287)
                    (1, 0/287)
                    (2, 1.5/287)
                    (3, 19/287)
                    (4, 67/287)
                    (5, 135.5/287)
                };
                \addplot[mctsendgame, mark=*] coordinates {
                    (0, 0)
                    (1, 0)
                    (2, 0)
                    (3, 0.08)
                    (4, 0.52)
                    (5, 1)
                };
                \addplot[ourmax, mark=*] coordinates {
                    (0, 0)
                    (1, 0)
                    (2, 0)
                    (3, 0.08)
                    (4, 0.52)
                    (5, 1)
                };
            }
            \coordinate (bot) at (rel axis cs:0,1);
        \end{groupplot}
        \path (top)--(bot) coordinate[midway] (group center);
        \node[below=-1pt,inner sep=0pt,anchor=north] at (current bounding box.south -| group center) {\pgfplotslegendfromname{zelda}};
    \end{tikzpicture}
}

\newcommand{\chainloop}{
    \begin{tikzpicture}
        \begin{groupplot}[
            group style={group size=4 by 1},
                ymin=-0.1,
                ymax=1.1,
                xmin=-0.5,
                xmax=5.5,
                grid=major,
                extra y ticks={0.25,0.75},
                xtick={0,...,5},
                xticklabels={5, 10, 25, 50, 100, 250},
        ]
            \nextgroupplot[
                title=ChainLoop-10,
                legend to name=link,
                ylabel=Return,
                xlabel=\nsims,
            ]{
                \addplot[mcts, mark=*] coordinates {
                    (0, 259.5/287)
                    (1, 283.5/287)
                    (2, 287/287)
                    (3, 287/287)
                    (4, 287/287)
                    (5, 287/287)
                };
                \addlegendentry{MCTS};
                \addplot[mctst+, mark=*] coordinates {
                    (0, 287/287)
                    (1, 287/287)
                    (2, 287/287)
                    (3, 287/287)
                    (4, 287/287)
                    (5, 287/287)
                };
                \addlegendentry{MCTS-T+};
                \addplot[mctsendgame, mark=*] coordinates {
                    (0, 1)
                    (1, 1)
                    (2, 1)
                    (3, 1)
                    (4, 1)
                    (5, 1)
                };
                \addlegendentry{\ours{} / \ourmax{} (ours)};
                \coordinate (top) at (rel axis cs:1,0);
            }
            \nextgroupplot[
                title=ChainLoop-25,
            ]{
                \addplot[mcts, mark=*] coordinates {
                    (0, 0/287)
                    (1, 0/287)
                    (2, 0/287)
                    (3, 0/287)
                    (4, 0/287)
                    (5, 7/287)
                };
                \addplot[mctst+, mark=*] coordinates {
                    (0, 0/287)
                    (1, 6.5/287)
                    (2, 287/287)
                    (3, 287/287)
                    (4, 287/287)
                    (5, 287/287)
                };
                \addplot[mctsendgame, mark=*] coordinates {
                    (0, 1)
                    (1, 1)
                    (2, 1)
                    (3, 1)
                    (4, 1)
                    (5, 1)
                };
            }
            \nextgroupplot[
                title=ChainLoop-50,
            ]{
                \addplot[mcts, mark=*] coordinates {
                    (0, 0/287)
                    (1, 0/287)
                    (2, 0/287)
                    (3, 0/287)
                    (4, 0/287)
                    (5, 0/287)
                };
                \addplot[mctst+, mark=*] coordinates {
                    (0, 0/287)
                    (1, 0/287)
                    (2, 2.5/287)
                    (3, 287/287)
                    (4, 287/287)
                    (5, 287/287)
                };
                \addplot[mctsendgame, mark=*] coordinates {
                    (0, 1)
                    (1, 1)
                    (2, 1)
                    (3, 1)
                    (4, 1)
                    (5, 1)
                };
            }
            \nextgroupplot[
                title=ChainLoop-100,
            ]{
                \addplot[mcts, mark=*] coordinates {
                    (0, 0/287)
                    (1, 0/287)
                    (2, 0/287)
                    (3, 0/287)
                    (4, 0/287)
                    (5, 0/287)
                };
                \addplot[mctst+, mark=*] coordinates {
                    (0, 0/287)
                    (1, 0/287)
                    (2, 0/287)
                    (3, 0/287)
                    (4, 287/287)
                    (5, 287/287)
                };
                \addplot[mctsendgame, mark=*] coordinates {
                    (0, 1)
                    (1, 1)
                    (2, 1)
                    (3, 1)
                    (4, 1)
                    (5, 1)
                };
            }
            \coordinate (bot) at (rel axis cs:0,1);
        \end{groupplot}
        \path (top)--(bot) coordinate[midway] (group center);
        \node[below=-1pt,inner sep=0pt,anchor=north] at (current bounding box.south -| group center) {\pgfplotslegendfromname{link}};
    \end{tikzpicture}
}

\newcommand{\frozenlakeSmall}{
    \begin{tikzpicture}
        \begin{axis}[
            title=FrozenLake,
            ymin=-0.1,
            ymax=1.1,
            xmin=-0.5,
            xmax=5.5,
            grid=major,
            extra y ticks={0.25,0.75},
            xtick={0,...,5},
            xticklabels={5, 10, 25, 50, 100, 250},
            ylabel=Return,
            xlabel=\nsims,
            legend to name=ganondorf1,
            legend style={
                at={(0.24,0.5)},
                anchor=north,
                legend columns=1,
            },
        ]
        \addplot[mcts, mark=*] coordinates {
            (0, 4/276)
            (1, 14/276)
            (2, 30.5/276)
            (3, 76.5/276)
            (4, 145.5/276)
            (5, 220.5/276)
        };
        \addlegendentry{MCTS};
        \addplot[mctst, mark=*] coordinates {
            (0, 8/276)
            (1, 19/276)
            (2, 40.5/276)
            (3, 86.5/276)
            (4, 171/276)
            (5, 251/276)
        };
        \addlegendentry{MCTS-T};
        \addplot[mctst+, mark=*] coordinates {
            (0, 4.5/276)
            (1, 17/276)
            (2, 72.5/276)
            (3, 153.5/276)
            (4, 237.5/276)
            (5, 274/276)
        };
        \addlegendentry{MCTS-T+};
        \addplot[mctsendgame, mark=*] coordinates {
            (0, 0.1154)
            (1, 0.67)
            (2, 0.84)
            (3, 0.96)
            (4, 1.0)
            (5, 1.0)
        };
        \addlegendentry{\ours{} (ours)};
        \addplot[ourmax, mark=*] coordinates {
            (0, 0.36)
            (1, 0.56)
            (2, 0.96)
            (3, 0.96)
            (4, 1)
            (5, 1)
        };
        \addlegendentry{\ourmax{} (ours)};
        \end{axis}
        \node[right=5pt,inner sep=0pt,anchor=west] at (current axis.east) {\pgfplotslegendfromname{ganondorf1}};
    \end{tikzpicture}
}

%% file: chapters/introduction.tex
\section{Introduction}%
\label{sec:intro}%
\begin{figure*}
    \centering
    \def\svgwidth{\linewidth}
    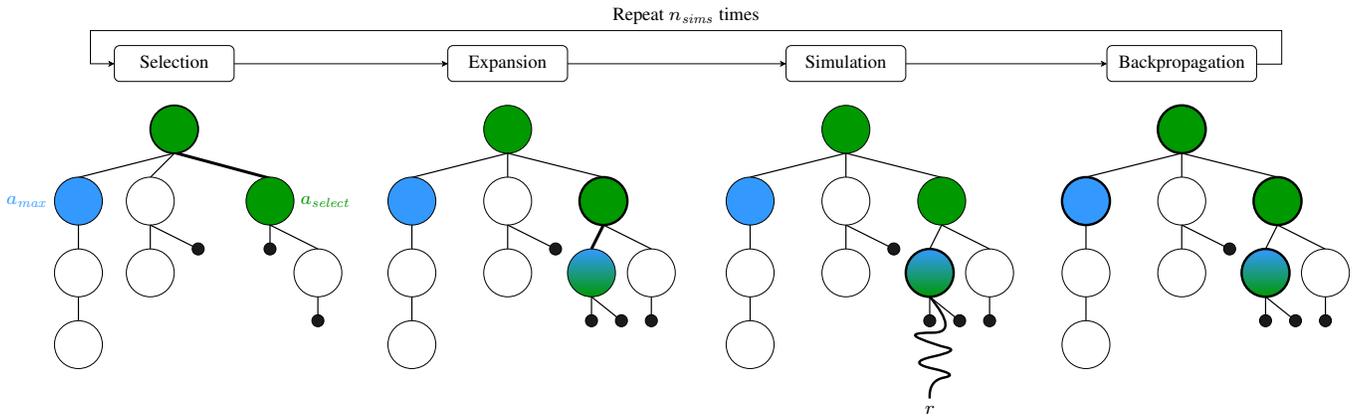
    \caption{\textbf{Improving MCTS by ignoring already explored subtrees and leaves by focusing on the unknown.} Updating the search strategy within MCTS by separating ``incrementing visit counts'' (displayed in blue) from the selected path (displayed in green) to explore more while keeping the number of iterations \(n_\mathit{sims}\) the same.}
    \label{fig:pipeline}
\end{figure*}

Monte-Carlo tree search (MCTS) is a powerful anytime best first search algorithm.
By utilizing the Upper Confidence Bounds (UCB) applied to trees (UCT)~\cite{kocsis_bandit_2006} as selection strategy to balance the exploration-exploitation trade-off, it achieves astonishing performance. Further the number of nodes it explores greatly affects its performance. However, especially in trees with a lot of terminal nodes, MCTS may explore the same nodes multiple times, without gaining new intel. This is a problem, wasting important resources. Thus, in this paper we propose an adaptation of MCTS, called \textit{\ours{}} which allows us to explore more nodes without additional computational budget by primarily adapting the selection strategy.

MCTS is frequently used to solve other reinforcement learning problems formalized as Markov decision processes (MDPs).
Within this broad class of problems, we focus on deterministic discrete action space MDPs. Although, this seems like a huge constraint, this problem class has has a lot of desired properties while yielding a great variety of interesting and important real world applications, e.g., truss design~\cite{luo_alphatruss_2022}, learning decision trees~\cite{nunes_monte_2018}, finding virtual network embeddings~\cite{zheng_single-player_2021}, predicting chemical molecular structure from infrared and NMR spectra~\cite{sridharan_spectra_2021,devata_deepspinn_2023}. \citeauthor{moerland_second_2020} \shortcite{moerland_second_2020} have identified that many of these single-player tasks have paths that lead to early termination which makes them perfectly fit for our adaptation.

MCTS recently showed great success, combined with neural networks, demonstrated by AlphaGo~\cite{silver2016mastering}, AlphaZero~\cite{silver_general_2018} and fine-tuning of large language models~\cite{kamienny_end--end_2022}.
However, the problem of exploring nodes multiple times remains here, which is even more severe when using MCTS in environments where computational budget plays an important role, e.g., competitive chess.
Consequently, those algorithms may also benefit from our efficiency improvements, i.e., during training and the real-time search at test time \cite{moerland_second_2020}. 
\newline

Following this motivation, our contributions can be summarized as:
\begin{itemize}
    \item We propose a solution for the exploration issue in MCTS to explore terminal states multiple times.
    \item We introduce \ours{} an improved MCTS adaptation that precludes the revisitation of fully explored subtrees in subsequent iterations by adapting the \qvalue{}s and visitation counts carefully.
    \item We therefore ensure that every MCTS-simulation leads to new information.
    \item \ours{} can be used without customization  and for any deterministic discrete action space MDP and is guaranteed to perform as well or better than classical MCTS.
\end{itemize}
We divide our work into five parts, starting off by introducing our modified MCTS algorithm \ours{} including some theoretical properties.
We then experimentally evaluate its performance and search space coverage.
Before concluding and considering possible future work, we also look at related work and touch on limitations. 

%% file: images/endgame_pipeline.pdf_tex
\begingroup%
  \makeatletter%
  \providecommand\color[2][]{%
    \errmessage{(Inkscape) Color is used for the text in Inkscape, but the package 'color.sty' is not loaded}%
    \renewcommand\color[2][]{}%
  }%
  \providecommand\transparent[1]{%
    \errmessage{(Inkscape) Transparency is used (non-zero) for the text in Inkscape, but the package 'transparent.sty' is not loaded}%
    \renewcommand\transparent[1]{}%
  }%
  \providecommand\rotatebox[2]{#2}%
  \newcommand*\fsize{\dimexpr\f@size pt\relax}%
  \newcommand*\lineheight[1]{\fontsize{\fsize}{#1\fsize}\selectfont}%
  \ifx\svgwidth\undefined%
    \setlength{\unitlength}{1675.5bp}%
    \ifx\svgscale\undefined%
      \relax%
    \else%
      \setlength{\unitlength}{\unitlength * \real{\svgscale}}%
    \fi%
  \else%
    \setlength{\unitlength}{\svgwidth}%
  \fi%
  \global\let\svgwidth\undefined%
  \global\let\svgscale\undefined%
  \makeatother%
  \begin{picture}(1,0.3133393)%
    \lineheight{1}%
    \setlength\tabcolsep{0pt}%
    \put(0,0){\includegraphics[width=\unitlength,page=1]{endgame_pipeline.pdf}}%
    \put(0.0268712,0.1642126){\color[rgb]{0.2,0.6,1}\makebox(0,0)[rt]{\lineheight{1.25}\smash{\begin{tabular}[t]{r}\scriptsize$a_\mathit{max}$\end{tabular}}}}%
    \put(0.21531445,0.16421144){\color[rgb]{0,0.6,0}\makebox(0,0)[lt]{\lineheight{1.25}\smash{\begin{tabular}[t]{l}\scriptsize $a_\mathit{select}$\end{tabular}}}}%
    \put(0.68598926,0.00783348){\color[rgb]{0,0,0}\makebox(0,0)[t]{\lineheight{1.25}\smash{\begin{tabular}[t]{c}\scriptsize $r$\end{tabular}}}}%
    \put(0.12108326,0.26648687){\color[rgb]{0,0,0}\makebox(0,0)[t]{\lineheight{1.25}\smash{\begin{tabular}[t]{c}\scriptsize Selection\end{tabular}}}}%
    \put(0.37041182,0.26648688){\color[rgb]{0,0,0}\makebox(0,0)[t]{\lineheight{1.25}\smash{\begin{tabular}[t]{c}\scriptsize Expansion\end{tabular}}}}%
    \put(0.6233214,0.26648688){\color[rgb]{0,0,0}\makebox(0,0)[t]{\lineheight{1.25}\smash{\begin{tabular}[t]{c}\scriptsize Simulation\end{tabular}}}}%
    \put(0.50370626,0.30214861){\color[rgb]{0,0,0}\makebox(0,0)[t]{\lineheight{1.25}\smash{\begin{tabular}[t]{c}\scriptsize Repeat $n_\mathit{sims}$ times\end{tabular}}}}%
    \put(0.87444047,0.26648688){\color[rgb]{0,0,0}\makebox(0,0)[t]{\lineheight{1.25}\smash{\begin{tabular}[t]{c}\scriptsize Backpropagation\end{tabular}}}}%
  \end{picture}%
\endgroup%

%% file: chapters/method.tex
\section{Method}
MCTS consists of four steps: selection, expansion, simulation and backpropergation, as visualized in \cref{fig:pipeline}. We will first briefly explain how these steps work in the classical MCTS and then go into detail with our adaptation of each step. 
\subsection{Classical MCTS}
Classical MCTS \cite{kocsis_bandit_2006,coulom_efficient_2007} starts with a \textit{selection phase}, wherein it traverses the tree from the root to a leaf node $l$, employing the UCT formula
\begin{equation}
    \textrm{UCT} = \dfrac{W_i}{N_c} + C \cdot \sqrt{\frac{\ln(N_p)}{N_c}}.
    \label{equ:uct}
\end{equation}
\(W\) represents the summed up reward of a specific node, \(C\) is a tuneable exploration parameter controlling the balance between exploration and exploitation. We stayed with the default $C = \sqrt{2}$. \(N_p\) denotes the total visit count of the parent node, and \(N_c\) signifies the visit count of the child node being evaluated. The formula combines these elements to estimate the upper confidence bound for a node's potential value, guiding the MCTS to strategically explore the decision space.
Subsequently, in the \textit{expansion step}, new nodes are added to the tree to represent potential future actions. The \textit{simulation phase} follows, wherein a randomized or heuristic-based rollout is executed from the selected node to estimate the potential outcome. In current implementations, starting with AlphaGo \cite{silver2016mastering}, the rollout is often replaced with an evaluator neural network to predict the quality of a state. The simulation path ends, returning a reward $r$.
The final step, \textit{backpropagation}, updates the statistics of the traversed nodes based on the simulation result, refining the algorithm's understanding of the decision space. 
\subsection{\ours{}}
\label{sec:MCTS_Endgame}
In this paper, we propose a novel MCTS formula, called \textbf{Am}plifed \textbf{Ex}ploration MCTS (\ours{})\footnote{\url{https://github.com/wwjbrugger/AmEx-MCTS}}, described in \cref{alg:MCTSEndgame}, improving the overall coverage of the search tree and increasing the relative performance while using the same computational budget as our baseline, MCTS.

For this work, we assume a deterministic MDP, as such there is neither randomness in the transition function $\mathcal{T}$ nor in the reward function $\mathcal{R}$. Hence, it is sufficient to visit each terminal state only once. However, in the classical MCTS algorithm there is no rule preventing visiting states multiple times, using valuable computational time on already known information. To prevent evaluating not only terminal states but subtrees multiple times, we adapt MCTS selection and update strategies to ignore fully explored parts of the search tree while keeping the exploration-exploitation mechanism and structure of MCTS mostly untouched. 

For this, we track every not completely explored and new subtrees/actions in a separate list of state-action pairs called \(nces\) (not completely explored subtree).
\bigskip

\input{pseudo_code/MCTS}

\begin{figure}[t]
    \centering
    \def\svgwidth{.8\linewidth}
    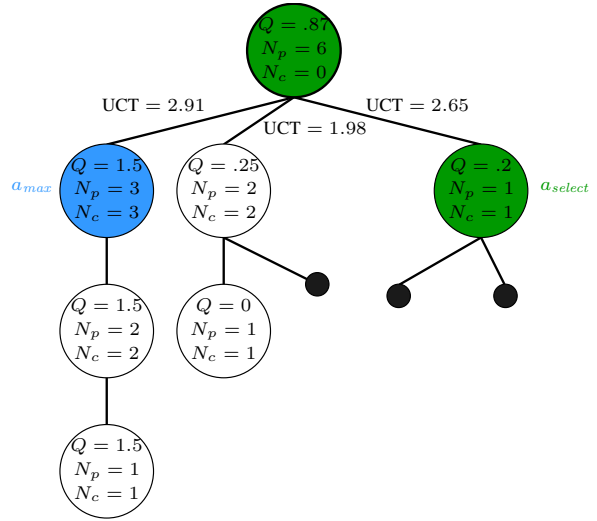
    \caption{\textbf{Ignoring fully explored subtrees within the search.} \ours{} introduces two new parameters \(a_\mathit{max}\) and \(a_\mathit{select}\) to differentiate between fully explored subtrees and such who are not. Selecting \(a_\mathit{select}\), ignores the already fully explored left subtree, while \(a_\mathit{max}\) would only lead in already known states. \(N_p\) describes how often the node was visited within the search, \(N_c\) the number how often this node had the highest UCT value among its siblings and \(Q\) the value of the state.}
    \label{fig:selection}
\end{figure}

In the following, we will revisit each phase within MCTS (cf. \cref{fig:pipeline}), describing the changes we made:

\paragraph{Selection} When choosing the next node in the selection process, MCTS follow the UCT equation (\autoref{equ:uct}), taking the action with the highest value. In this work, these actions are called \(a_\mathit{max}\). However, we make a distinction here and introduce \(a_\mathit{select}\), defined as action with the highest value of all actions which have not been completely explored yet, i.e., the subtree is not complete and the action is still listed in $\mathit{nces}[s, \placeholder]$. 
Opposed to MCTS, we use $a_\mathit{select}$ to explore the search tree, however, when updating the visit count in the backpropagation step, the visit count of the edge associated with $a_\mathit{max}$ is incremented according to MCTS, so we need to track these actions too. This process can be seen in Figure~\ref{fig:selection} and is also described in Algorithm~\ref{alg:search}. In large search trees with a lot of unknown nodes, our approach behaves like MCTS since in such situation both approaches explore new nodes per iteration and fully explored subtress are rare, occuring only near terminal states or in shallow trees. However, in these cases \ours{} prevents going back into a already fully explored part of the search tree unnecessarily and instead explores new parts. 

\input{pseudo_code/search}

\paragraph{Expansion} In MCTS, each node is defined by its state representation, that we call $s_{\text{MDP}}$, and its parent node. If we can reach a state over multiple paths in the tree, we know that the subtree behind these nodes are the same as behind any other node with the same state representation. This is known as transpositions in the MCTS community. As such, we do not have to reevaluate each subtree separately. When extending the tree, we calculate a hash-key of $s_{\text{MDP}}$ to check whether the state representation already occurred somewhere else in the tree. If so, we do not simulate but instead go directly to the backpropagation phase and mark the state as terminal with a reward equal to the current \qvalue{} of the identical state representation. This extension can be seen in \cref{alg:rollout}.

\input{pseudo_code/rollout}

\paragraph{Simulation} The simulation phase stays untouched by us. In \cref{alg:rollout} \cref{lst:line:sim}, the function \textsc{Simulate} can either be a neural network call or it can be a random rollout or a policy driven rollout to get the a reward $r$. For $\mathcal{R}$, we assume $\mathcal{R} \in \mathbb{R}$. 

\begin{figure}[t]
    \centering
    \def\svgwidth{.8\linewidth}
    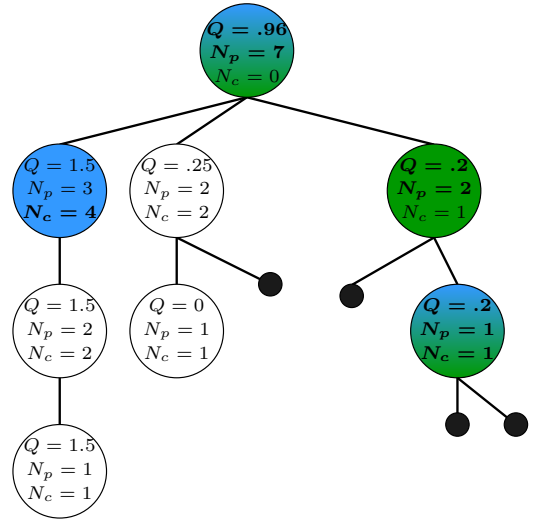
    \caption{\textbf{Updating the visit counts $\boldsymbol{N_c}$ along the original MCTS, to keep them representative for the output policy.} Within the backpropagation step of MCTS we separate the visit count updated along the selected path (green) and the visit counts which would be selected by the original MCTS algorithm (blue). The \qvalue{}s are updated along the selected path.}
    \label{fig:backprop}
\end{figure}

\paragraph{Backpropagation} When backpropagating in MCTS, the idea is to increase the visit counts of all nodes in the selected path as well as adding the reward $r$ to the summed up reward of each of these nodes to update the UCT values for the next selection phase. This is the major step to keep a balance between exploration and exploitation. In case of \ours{}, this is not as easily possible since we distinguish between moves MCTS would have made (\cref{fig:backprop} blue path) and the actual path we selected (\cref{fig:backprop} green path). Since we want to keep the exploration and exploitation balance in check and only add to the exploration in case of unneeded selection of already known nodes, the goal is not to break any theoretical guarantees of the UCT algorithm. To achieve this, some additional steps are necessary in the backpropagation phase. 
First, we needed to not only distinguish between \(a_\mathit{max}\) and \(a_\mathit{select}\) but also between \(N_p\) and \(N_c\). While \(N_c\) is the number of visits the node would have gotten following MCTS, \(N_p\) is the number of real selections. Updating them separately is important so that in the end, when creating the policy (cf. \cref{alg:MCTSEndgame} \cref{alg:line:policycreation}) we can still rely on the visit counts of nodes. Important is that we only adapt \(N_c\) within the blue path but have to prevent the \qvalue{} from changing, so we also have to update the accumulated reward \(W\) such that \(Q\) stays the same. In general, when updating the accumulated rewards, we also have to distinguish between two cases. First the easy case, if \(a_\mathit{max}\) is equal to \(a_\mathit{select}\), in this case we can propagate the reward as normal, adding it to the accumulated one and are done. 

However, to stay as close to the classical MCTS as possible, we only backpropagate the reward $r$ found with $a_{select}$ if it is greater than what would be found from following $a_{max}$. As a result, we do not propagate bad moves more than MCTS would.

Therefore, we set $r$ to the expected outcome for following $a_\mathit{max}$. In addition, by not revisiting fully explored subtrees, we prevent the \qvalue{} ($\nicefrac{s.W}{s.N_c}$) to converge to the true value $q_\pi$. Thus, we manually set it to the maximum of the children's \qvalue{}s and even speed up this process.
It is also to notice that, in contrast to the standard MCTS formulation, we discount the reward using the factor $\gamma$ as commonly done for MDPs in reinforcement learning.

\input{pseudo_code/backup}

\subsection{Properties}
Before showing empirical results, we want to briefly present some theoretical properties of our adaptations.

While keeping the changes to the original MCTS algorithm minimal, in the limit, \ours{} converges to an exhaustive search and therefore always returns the true values given enough time.

Further, the original asymptotic guarantees \cite{kocsis_bandit_2006} of the underlying UCT and MCTS still hold true as discussed in the following.
We can distinguish three cases:
\begin{enumerate}
    \item No terminal states have been encountered so far
    \item The search tree has been fully explored
    \item A mixture of the above cases
\end{enumerate}
In the first case, the MCTS does not change at all, i.e., all guarantees and properties are preserved.
In the second case, \ours{} does not run any simulations at all anymore and takes the optimal actions leading to the highest outcome (cf. \cref{alg:MCTSEndgame} \cref{lst:line:full}).

In the last case, we can focus on different parts of the MCTS:
\begin{enumerate}
    \item The returned policy $\boldsymbol{\pi}$
    \begin{enumerate}
        \item When terminal nodes have only been found directly below the root
        \item When terminal nodes have been found anywhere
    \end{enumerate}
    \item The backpropagated value $r$
\end{enumerate}
In case where there are only terminal nodes below the root, the visit counts $N_c$ are updated according to what classical MCTS would do, as motivated earlier, and therefore, the returned policy $\boldsymbol{\pi}$ is the same as in classical MCTS.

The last case is the most interesting. To discover the behavior, we need to look at the influence of our adaptations which boil down to the UCT score. The values in UCT that are affected by \ours{} are the \qvalue{}s and the visit counts $N_p$ and $N_c$, respectively.

When considering the \qvalue{}s, we can pick a random node that has a fully explored subtree among the child nodes. Here, \ours{} mostly just produces a \qvalue{} similar to classical MCTS but since more nodes are expanded and more simulations are run, the estimate is very likely to converge much faster and be more precise.

This now leaves us with the visit counts. For selecting an action according to the UCT formula (cf. \autoref{equ:uct}), we only need the $N_p$ of the current state and the $N_c$ of the children.
In addition, the path to the current node does not matter.
Hence, we can neglect the fact that the subtree in question has been visited more often than in in the classical MCTS and the resulting values $N_p$ and $N_c$ only compose the result of the same search with a bigger $n_\mathit{sims}$.
This also implies that it also does not matter that the subtree grew bigger than in MCTS.
\bigskip

On that account, we have shown that \textbf{the estimate of the \qvalue{}s will be at least as good as with classical MCTS}, the visit counts $N_p$ are higher than in classical MCTS but match with the intention of it and the $N_c$ are aligned with what classical MCTS would have. Consequently, \textit{the convergence properties of MCTS with UCT still hold}.

\subsection{\ourmax{}: Replacing the Mean with the Max}
Normally, by using the average reward, the algorithm considers the overall performance of an action, smoothing out fluctuations that may arise from limited or noisy samples. 
Depending on the problem description, however, we may not want to select the move with the best potential, as is done in the definition of UCT \cite{kocsis_bandit_2006}, but simply want to choose the action that has performed best in our previous search. Especially in situations other than two-player zero-sum games, this can lead to very good results. 
\citeauthor{petersen_deep_2021} \shortcite{petersen_deep_2021} employ a risk-seeking policy gradient in the domain of equation discovery by selecting only the best $(1 - \varepsilon)$-quantile of the reward to update their network. We adopt this idea for \ourmax{} by substitution the mean function by the maximum function, i.e., only using the 100\%-quantile, within the backpropagating of the \qvalue{}s in our search tree.
Therefore, in such cases, we replace $\nicefrac{W_i}{N_c}$ with
\begin{equation}\label{equ:updateQsa}
    Q(s,a) \gets \max\big(Q(s,a), r\big)
\end{equation}
within the UCT formula (see \autoref{equ:uct}).

%% file: pseudo_code/MCTS.tex
\begin{algorithm}[tb]
\caption{\textsc{\ours{}}}\label{alg:MCTSEndgame}
    \begin{algorithmic}[1]
        \Require $\text{state } s, n_\mathit{sims}$
        \State create root node $\mathit{root}$ from state $s$ \;
        \State $\mathit{nces} \gets \mathcal{T}(root,\cdot)$ 
        \Comment{List of not fully explored subtrees}
        \State $t \gets$ \{$\mathit{root}$\}\; \Comment{Initialize tree $t$ with root node}
        \State $S \gets \varnothing $  \Comment{Initialize set of visited states}
        \For{$i \gets 1$ to $n_\mathit{sims}$}
        \State $l \gets \textsc{Select}(\mathit{root})$ \Comment{Select leaf node $l$}
        \State $r,t \gets \textsc{ExpandAndSimulate}(l, t)$
        \State $t \gets \textsc{Backpropagate}(l,r)$
        \If{$t$ is completely explored}
        \State $Q(root,\placeholder) \gets \nicefrac{\mathcal{T}(root, \placeholder).W}{\mathcal{T}(root, \placeholder).N_p}$
        \State \Return Create policy $\boldsymbol{\pi}$ using $Q(root,\placeholder)$ \label{lst:line:full}
        \label{lst:line:probs}
        \EndIf
        \EndFor
        \State \Return Create policy $\boldsymbol{\pi}$ from tree $t$ \label{alg:line:policycreation}
    \end{algorithmic}
\end{algorithm}

%% file: images/Selection2.pdf_tex
\begingroup%
  \makeatletter%
  \providecommand\color[2][]{%
    \errmessage{(Inkscape) Color is used for the text in Inkscape, but the package 'color.sty' is not loaded}%
    \renewcommand\color[2][]{}%
  }%
  \providecommand\transparent[1]{%
    \errmessage{(Inkscape) Transparency is used (non-zero) for the text in Inkscape, but the package 'transparent.sty' is not loaded}%
    \renewcommand\transparent[1]{}%
  }%
  \providecommand\rotatebox[2]{#2}%
  \newcommand*\fsize{\dimexpr\f@size pt\relax}%
  \newcommand*\lineheight[1]{\fontsize{\fsize}{#1\fsize}\selectfont}%
  \ifx\svgwidth\undefined%
    \setlength{\unitlength}{330.75bp}%
    \ifx\svgscale\undefined%
      \relax%
    \else%
      \setlength{\unitlength}{\unitlength * \real{\svgscale}}%
    \fi%
  \else%
    \setlength{\unitlength}{\svgwidth}%
  \fi%
  \global\let\svgwidth\undefined%
  \global\let\svgscale\undefined%
  \makeatother%
  \begin{picture}(1,1)%
    \lineheight{1}%
    \setlength\tabcolsep{0pt}%
    \put(0,0){\includegraphics[width=\unitlength,page=1]{Selection2.pdf}}%
    \put(0.45451898,0.94707629){\color[rgb]{0,0,0}\makebox(0,0)[t]{\lineheight{0.89999998}\smash{\begin{tabular}[t]{c}\scriptsize$Q = .87$\\\scriptsize$N_p = 6$\\\scriptsize$N_c = 0$\end{tabular}}}}%
    \put(0.09170719,0.67496745){\color[rgb]{0,0,0}\makebox(0,0)[t]{\lineheight{0.89999998}\smash{\begin{tabular}[t]{c}\scriptsize$Q = 1.5$\\\scriptsize$N_p = 3$\\\scriptsize$N_c = 3$\end{tabular}}}}%
    \put(0.31846456,0.67496745){\color[rgb]{0,0,0}\makebox(0,0)[t]{\lineheight{0.89999998}\smash{\begin{tabular}[t]{c}\scriptsize$Q = .25$\\\scriptsize$N_p = 2$\\\scriptsize$N_c = 2$\end{tabular}}}}%
    \put(0.31846456,0.40285862){\color[rgb]{0,0,0}\makebox(0,0)[t]{\lineheight{0.89999998}\smash{\begin{tabular}[t]{c}\scriptsize$Q = 0$\\\scriptsize$N_p = 1$\\\scriptsize$N_c = 1$\end{tabular}}}}%
    \put(0.09170719,0.40285862){\color[rgb]{0,0,0}\makebox(0,0)[t]{\lineheight{0.89999998}\smash{\begin{tabular}[t]{c}\scriptsize$Q = 1.5$\\\scriptsize$N_p = 2$\\\scriptsize$N_c = 2$\end{tabular}}}}%
    \put(0.09170719,0.13074977){\color[rgb]{0,0,0}\makebox(0,0)[t]{\lineheight{0.89999998}\smash{\begin{tabular}[t]{c}\scriptsize$Q = 1.5$\\\scriptsize$N_p = 1$\\\scriptsize$N_c = 1$\end{tabular}}}}%
    \put(0.81733077,0.67496745){\color[rgb]{0,0,0}\makebox(0,0)[t]{\lineheight{0.89999998}\smash{\begin{tabular}[t]{c}\scriptsize$Q = .2$\\\scriptsize$N_p = 1$\\\scriptsize$N_c = 1$\end{tabular}}}}%
    \put(0.93201687,0.63822909){\color[rgb]{0,0.6,0}\makebox(0,0)[lt]{\smash{\begin{tabular}[t]{l}\scriptsize$a_\mathit{select}$\end{tabular}}}}%
    \put(-0.01133693,0.63822909){\color[rgb]{0.2,0.6,1}\makebox(0,0)[rt]{\smash{\begin{tabular}[t]{r}\scriptsize$a_\mathit{max}$\end{tabular}}}}%
    \put(0.28293885,0.79113008){\color[rgb]{0,0,0}\makebox(0,0)[rt]{\smash{\begin{tabular}[t]{r}\scriptsize$\text{UCT} = 2.91$\end{tabular}}}}%
    \put(0.59443336,0.79113008){\color[rgb]{0,0,0}\makebox(0,0)[lt]{\smash{\begin{tabular}[t]{l}\scriptsize$\text{UCT} = 2.65$\end{tabular}}}}%
    \put(0.39602976,0.74540477){\color[rgb]{0,0,0}\makebox(0,0)[lt]{\smash{\begin{tabular}[t]{l}\scriptsize$\text{UCT} = 1.98$\end{tabular}}}}%
  \end{picture}%
\endgroup%

%% file: pseudo_code/search.tex
\begin{algorithm}[tb]
\caption{\textsc{Select}}\label{alg:search}
    \begin{algorithmic}[1]
        \Require $\text{state } s$
        \State $a_\mathit{select} \gets \argmax_{a \in \mathit{nces}} \textrm{UCT}(s,a)$
        \Comment{Select}
        \State $a_\mathit{max} \gets \argmax_{a \in \mathcal{A}(s)} \textrm{UCT}(s,a)$ 
        \State $s' \gets \mathcal{T}(s,a_\mathit{select})$
        \If{$s' \notin t \textbf{ or } s'.\mathit{terminal}$}
        \State \Return $s'$
        \Comment{Return the leaf node}
        \EndIf
        \State \Return \textsc{Select}($s'$) 
        \Comment{Traverse to leaf}
    \end{algorithmic}
\end{algorithm}

%% file: pseudo_code/rollout.tex
\begin{algorithm}[tb]
\caption{\textsc{ExpandAndSimulate}}\label{alg:rollout}
    \begin{algorithmic}[1]
        \Require $\text{state } s, \text{tree } t$
        \State Add $s$ to $t$ 
        \If {$s.s_\text{MDP} \in S'$} \Comment{Check if MDP state is known}
        \State $r \gets S'[s]$ \Comment{Use stored $r$}\;
        \State  $s.\mathit{terminal} \gets 1$
        \State \Return $r, t$
        \EndIf
        \If{$\mathcal{A}(s) = \varnothing$}
        \State $s.\mathit{terminal} \gets 1$
        \State \Return $\mathcal{R}(s)$
        \EndIf
        \State Add $\mathcal{T}(s,\placeholder)$ to $\mathit{nces}$
        \State \Return \textsc{Simulate}($s$), t \label{lst:line:sim}
        \Comment{Simulate}
    \end{algorithmic}
\end{algorithm}

%% file: pseudo_code/backup.tex
\algnewcommand\algorithmicforeach{\textbf{for each}}
\algdef{S}[FOR]{ForEach}[1]{\algorithmicforeach\ #1\ \algorithmicdo}

\begin{algorithm}[tb]
\caption{\textsc{Backpropagate}}\label{alg:backup}
    \begin{algorithmic}[1]
        \Require $\text{state } s, \text{reward } r$
        \If{$\lnot \, s.\mathit{terminal}$}
        \State $c_s \gets \mathcal{T}(s, a_\mathit{select})$
        \State $c_m \gets \mathcal{T}(s, a_\mathit{max})$
        \If{$ c_m \neq c_a \textbf{ and } r < \nicefrac{c_m.W}{c_m.N_c}$}
        \State $r \gets \nicefrac{c_m.W}{c_m.N_c}$ \Comment{Do not worsen average of the parent by extra exploration}
        \EndIf
        \State $c_s.N_p \gets c_s.N_p + 1$
        \State $c_m.N_c \gets c_m.N_c + 1 $
        \State $c_m.W \gets \nicefrac{c_m.W \cdot\ c_m.N_c }{c_m.N - 1}$
        \State $S'[c_s.s_\text{MDP}] \gets \nicefrac{c_s.W}{c_s.N_c}$
        
        \EndIf
        
        \State $s.W \gets s.W + r$
        
        \If{$c_s.\mathit{terminal}$}\Comment{Update $\mathit{nces}$}
        \State Remove $c_s$ from $\mathit{nces}$
        \If{ all $\mathcal{T}(s,\placeholder)$ were removed from $\mathit{nces}$}
        \State $s.\mathit{terminal} \gets 1$
        \State $s.W = \gamma \cdot s.N_c \cdot \; \max_{a \in \mathcal{A}(s)}\left(\nicefrac{\mathcal{T}(s, a).W}{\mathcal{T}(s, a).N_c}\right)$
        \EndIf
        \EndIf
        \If{$s = \mathit{root}$}
        \State $s.N_p \gets s.N_p + 1 $
        \State \Return $t$
        \EndIf
        \State \textsc{Backpropagate}$(s.\mathit{parent}, \gamma \cdot r)$
    \end{algorithmic}
\end{algorithm}

%% file: images/Backprop_latex.pdf_tex
\begingroup%
  \makeatletter%
  \providecommand\color[2][]{%
    \errmessage{(Inkscape) Color is used for the text in Inkscape, but the package 'color.sty' is not loaded}%
    \renewcommand\color[2][]{}%
  }%
  \providecommand\transparent[1]{%
    \errmessage{(Inkscape) Transparency is used (non-zero) for the text in Inkscape, but the package 'transparent.sty' is not loaded}%
    \renewcommand\transparent[1]{}%
  }%
  \providecommand\rotatebox[2]{#2}%
  \newcommand*\fsize{\dimexpr\f@size pt\relax}%
  \newcommand*\lineheight[1]{\fontsize{\fsize}{#1\fsize}\selectfont}%
  \ifx\svgwidth\undefined%
    \setlength{\unitlength}{330.75bp}%
    \ifx\svgscale\undefined%
      \relax%
    \else%
      \setlength{\unitlength}{\unitlength * \real{\svgscale}}%
    \fi%
  \else%
    \setlength{\unitlength}{\svgwidth}%
  \fi%
  \global\let\svgwidth\undefined%
  \global\let\svgscale\undefined%
  \makeatother%
  \begin{picture}(1,1)%
    \lineheight{1}%
    \setlength\tabcolsep{0pt}%
    \put(0,0){\includegraphics[width=\unitlength,page=1]{Backprop_latex.pdf}}%
    \put(0.45464853,0.93894845){\color[rgb]{0,0,0}\makebox(0,0)[t]{\lineheight{0.89999998}\smash{\begin{tabular}[t]{c}\scriptsize$\boldsymbol{Q = .96}$\\\scriptsize$\boldsymbol{N_p = 7}$\\\scriptsize$N_c = 0$\end{tabular}}}}%
    \put(0.09170719,0.67496745){\color[rgb]{0,0,0}\makebox(0,0)[t]{\lineheight{0.89999998}\smash{\begin{tabular}[t]{c}\scriptsize$Q = 1.5$\\\scriptsize$N_p = 3$\\\scriptsize$\boldsymbol{N_c = 4}$\end{tabular}}}}%
    \put(0.31846456,0.67496745){\color[rgb]{0,0,0}\makebox(0,0)[t]{\lineheight{0.89999998}\smash{\begin{tabular}[t]{c}\scriptsize$Q = .25$\\\scriptsize$N_p = 2$\\\scriptsize$N_c = 2$\end{tabular}}}}%
    \put(0.86268224,0.40285862){\color[rgb]{0,0,0}\makebox(0,0)[t]{\lineheight{0.89999998}\smash{\begin{tabular}[t]{c}\scriptsize$\boldsymbol{Q = .2}$\\\scriptsize$\boldsymbol{N_p = 1}$\\\scriptsize$\boldsymbol{N_c = 1}$\end{tabular}}}}%
    \put(0.31846456,0.40285862){\color[rgb]{0,0,0}\makebox(0,0)[t]{\lineheight{0.89999998}\smash{\begin{tabular}[t]{c}\scriptsize$Q = 0$\\\scriptsize$N_p = 1$\\\scriptsize$N_c = 1$\end{tabular}}}}%
    \put(0.09170719,0.40285862){\color[rgb]{0,0,0}\makebox(0,0)[t]{\lineheight{0.89999998}\smash{\begin{tabular}[t]{c}\scriptsize$Q = 1.5$\\\scriptsize$N_p = 2$\\\scriptsize$N_c = 2$\end{tabular}}}}%
    \put(0.09170719,0.13074977){\color[rgb]{0,0,0}\makebox(0,0)[t]{\lineheight{0.89999998}\smash{\begin{tabular}[t]{c}\scriptsize$Q = 1.5$\\\scriptsize$N_p = 1$\\\scriptsize$N_c = 1$\end{tabular}}}}%
    \put(0.81733077,0.67496745){\color[rgb]{0,0,0}\makebox(0,0)[t]{\lineheight{0.89999998}\smash{\begin{tabular}[t]{c}\scriptsize$\boldsymbol{Q = .2}$\\\scriptsize$\boldsymbol{N_p = 2}$\\\scriptsize$N_c = 1$\end{tabular}}}}%
  \end{picture}%
\endgroup%

%% file: chapters/experiments.tex
\section{Experiments}
\label{sec:experiments}
The primary goal of MCTS is to efficiently explore and evaluate decision spaces by iteratively building and sampling a search tree. The algorithm strives to achieve optimal performance in evaluating and selecting actions within a given environment.
After a short introduction of the domains, used in this work, we evaluate the performance of our methods, \ours\ and \ourmax, comparing it to the classical MCTS and MCTS-T~\cite{moerland_monte_2018}, an improvement with similar goals to \ours. In the last part, we compare the coverage of the search tree with the classical MCTS and highlight the effectiveness of our improvements in this aspect.

\subsection{Baseline Domains}
To evaluate and compare our method, we chose to use the same three domains that were already used by \citeauthor{moerland_monte_2018} \shortcite{moerland_monte_2018}, i.e., the Chain environment, the ChainLoop environment and a deterministic version of FrozenLake. These environments highlight problems within MCTS, regarding finding optimal actions. Thus, classical MCTS is not able to reach high rewards.

\paragraph{The Chain Environment}
The Chain environment is a task where the agent has to choose the correct action $k$ times in a row. At every step, the agent has $|\mathcal{A}(s)| = 2$ possible actions to choose from and receives a reward of 0. If the agent takes the wrong action, the episode terminates with a reward of 0. In case the agent finds the correct action in every state, it receives a reward of 1 and the episode terminates as well.

\paragraph{The ChainLoop Environment}
The ChainLoop environment is very similar to the Chain environment and only differs in one aspect, that is, it does not terminate when the agent chooses the wrong action but the environment transitions to the initial state $s_0$.

\paragraph{The Deterministic FrozenLake Environment}
The deterministic FrozenLake environment is a non-slippery version of the classical FrozenLake gym environment \cite{brockman2016openai}, this means that the environment always transitions into the next state according to the intended action. We use the $8 \times 8$ map and allow $T = 400$ time steps until the episode terminates.
\subsection{Comparison of the Performance}
\begin{figure}[t]
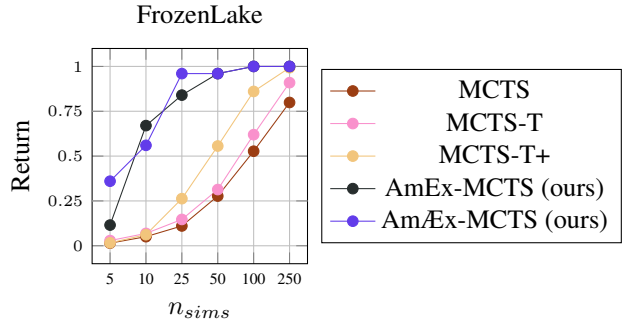

    \centering
    \frozenlakeSmall
    \caption{\textbf{Our approach dominates on the deterministic FrozenLake environment} as proposed by \protect\cite{moerland_monte_2018}.
    Higher values are better, where $1$ is the maximal return possible. The results reported are an average of 25 random seeds. Baseline results are taken from \protect\cite{moerland_monte_2018}.}
    \label{fig:frozenlake}
\end{figure}

\begin{figure*}[ht!]
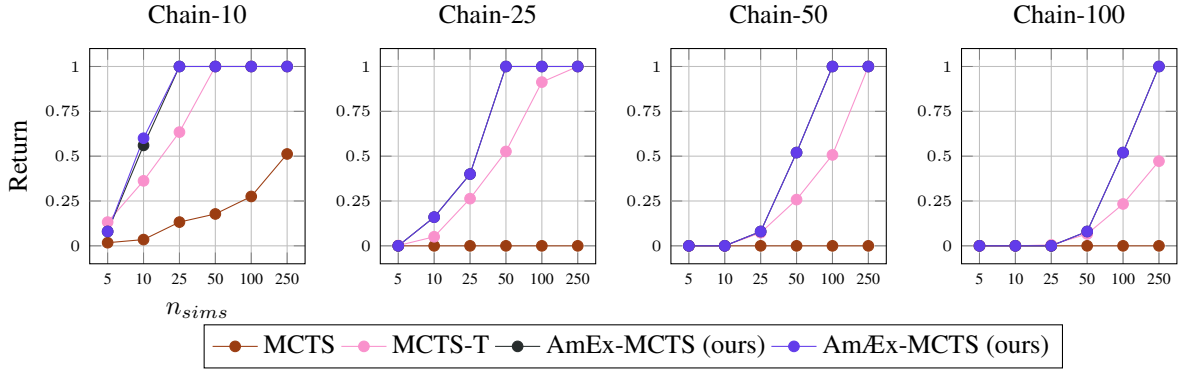

    \centering
    \chain
    \caption{\textbf{Our approach strongly outperforms the baselines on the Chain environment} as used in \protect\cite{moerland_monte_2018}.
    Higher values are better, where $1$ is the maximal return possible. The results reported are an average of 25 random seeds. Baseline results are taken from \protect\cite{moerland_monte_2018}.}
    \label{fig:chain-env}
\end{figure*}

\begin{figure*}[ht!]
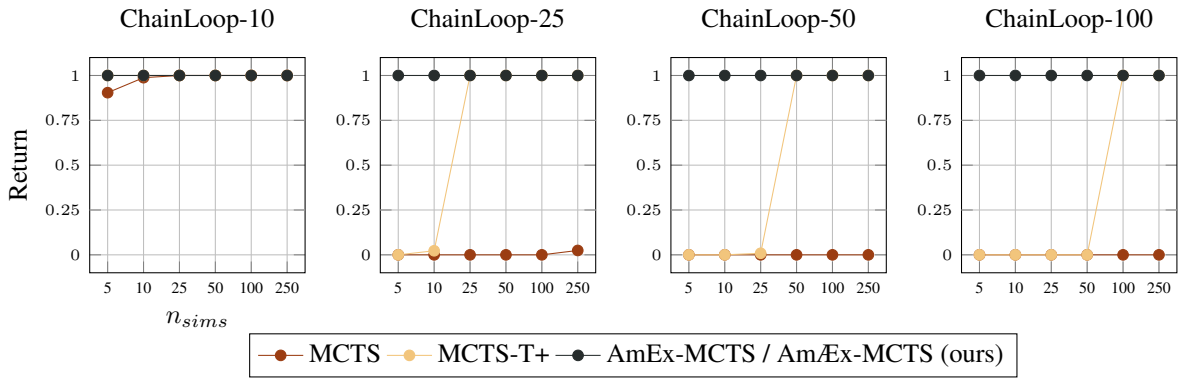

    \centering
    \chainloop
    \caption{\textbf{\ours{} achieves optimal results on the ChainLoop environment} as used in \protect\cite{moerland_monte_2018}.
    Higher values are better, where $1$ is the maximal return possible. The results reported are an average of 25 random seeds. Baseline results are taken from \protect\cite{moerland_monte_2018}.}
    \label{fig:chainloop-env}
\end{figure*}

In the experimental evaluation of our augmentation, we conducted comparisons between \ours{} and \ourmax{} against classical MCTS, as well as the closely related baselines MCTS-T and MCTS-T+ \cite{moerland_monte_2018}. The effectiveness of our enhancements is validated in the FrozenLake environment, as illustrated in \Cref{fig:frozenlake}, with \ours{} and \ourmax{} particularly excelling in scenarios with limited $n_{sims}$. 
The superiority of our approach is even more clear for the Chain and ChainLoop environments, as depicted in \Cref{fig:chain-env,fig:chainloop-env}. Notably, in the Chain environment, both \ours{} and \ourmax{} consistently outperform MCTS-T across all settings, showcasing optimal behavior in the ChainLoop environment. Considering the compelling evidence, we assert that the results effectively communicate the noteworthy advancements achieved through our approach.

Upon comparing \ours{} with \ourmax{}, it is observed that both agents exhibit comparable performance, with no clear advantage discernible between them. Subsequent comprehensive evaluation is needed to recommend one of them over the other. Presently, both \(Q\) formulas (mean and max) demonstrate validity within the context of MCTS for single-player tasks.

\subsection{Coverage of the Search Space}
\label{sec:cov}

\begin{figure*}[tbh]
    \centering
    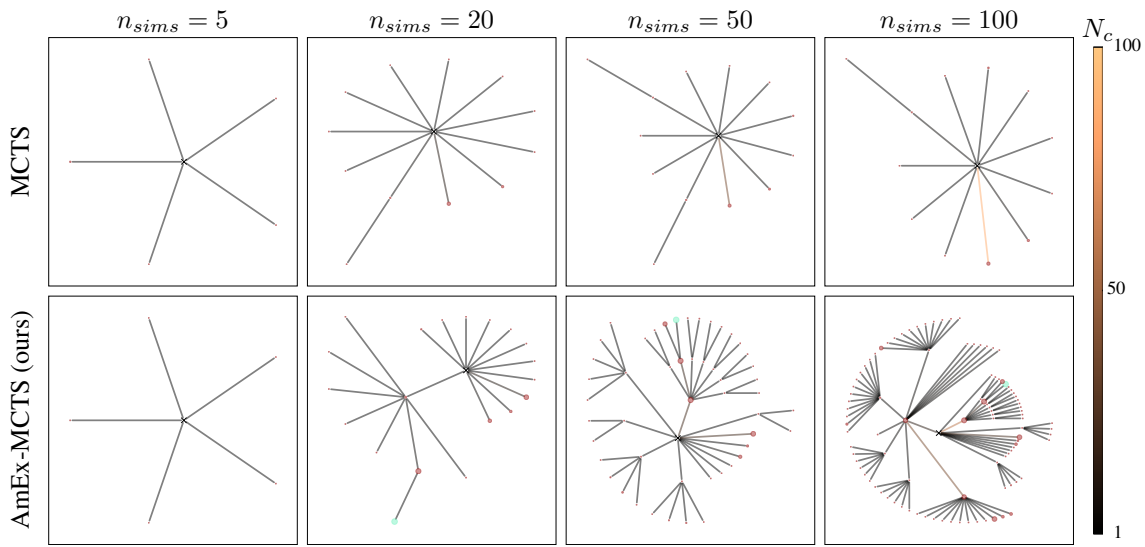
    \caption{\textbf{\ours{} results in a significantly broader exploration compared to classical MCTS.}
    The start node displayed as a cross. The size of the red nodes is proportional to the \qvalue{} after $n_{sims}$ steps. It is apparent that \ours{} covers a significant larger portion of the game tree and in contrast to classical MCTS and is able to find the optimal solution (green circle).}
    \label{fig:cov}
\end{figure*}

In addition to the performance results discussed in the previous section, we also report the coverage of the search trees (cf. \cref{fig:cov}).
The task is to sequentially generate a function $f_\mathit{gen}$ that describes the data $\mathcal{D} = (\boldsymbol{X}, \boldsymbol{y})$ given the rules from a grammar $\mathfrak{G}$  (see \cref{sec:grammar}).
The reward $\mathcal{R}(s)$ is $-1$ for incomplete or overly long equations or $1$ minus the mean square error for a completed equations.
The formula $f_\mathit{true}$ to create the data set for this experiment is $y = \sqrt{x_0}$ and can be created within three steps.
Even for higher simulation counts, classical MCTS keeps revisiting the same states and actions (visualized by brighter edge color) as for $n_\mathit{sims} = 20$ over and over again and gets stuck in a local optimum.
On the other hand, \ours{} uses every simulation step to explore an unknown path and finds the optimal solution (green circle) within less than $20$ simulations.

%% file: images/Comparision_searchtree_MCTS_Vanilla_Endgame.pdf_tex
\begingroup%
  \makeatletter%
  \providecommand\color[2][]{%
    \errmessage{(Inkscape) Color is used for the text in Inkscape, but the package 'color.sty' is not loaded}%
    \renewcommand\color[2][]{}%
  }%
  \providecommand\transparent[1]{%
    \errmessage{(Inkscape) Transparency is used (non-zero) for the text in Inkscape, but the package 'transparent.sty' is not loaded}%
    \renewcommand\transparent[1]{}%
  }%
  \providecommand\rotatebox[2]{#2}%
  \newcommand*\fsize{\dimexpr\f@size pt\relax}%
  \newcommand*\lineheight[1]{\fontsize{\fsize}{#1\fsize}\selectfont}%
  \ifx\svgwidth\undefined%
    \setlength{\unitlength}{428.39998627bp}%
    \ifx\svgscale\undefined%
      \relax%
    \else%
      \setlength{\unitlength}{\unitlength * \real{\svgscale}}%
    \fi%
  \else%
    \setlength{\unitlength}{\svgwidth}%
  \fi%
  \global\let\svgwidth\undefined%
  \global\let\svgscale\undefined%
  \makeatother%
  \begin{picture}(1,0.47226891)%
    \lineheight{1}%
    \setlength\tabcolsep{0pt}%
    \put(0,0){\includegraphics[width=\unitlength,page=1]{Comparision_searchtree_MCTS_Vanilla_Endgame.pdf}}%
    \put(0.97016098,0.4355883){\color[rgb]{0,0,0}\makebox(0,0)[lt]{\lineheight{1.25}\smash{\begin{tabular}[t]{l}\scriptsize 100\end{tabular}}}}%
    \put(0.96447121,0.22131901){\color[rgb]{0,0,0}\makebox(0,0)[lt]{\lineheight{1.25}\smash{\begin{tabular}[t]{l}\scriptsize 50\end{tabular}}}}%
    \put(0.97016098,0.00666675){\color[rgb]{0,0,0}\makebox(0,0)[lt]{\lineheight{1.25}\smash{\begin{tabular}[t]{l}\scriptsize 1\end{tabular}}}}%
    \put(0.01596146,0.33839369){\color[rgb]{0,0,0}\rotatebox{90}{\makebox(0,0)[t]{\lineheight{1.25}\smash{\begin{tabular}[t]{c}MCTS\end{tabular}}}}}%
    \put(0.01596146,0.11089328){\color[rgb]{0,0,0}\rotatebox{90}{\makebox(0,0)[t]{\lineheight{1.25}\smash{\begin{tabular}[t]{c}\ours{} (ours)\end{tabular}}}}}%
    \put(0.14279482,0.45610229){\color[rgb]{0,0,0}\makebox(0,0)[t]{\lineheight{1.25}\smash{\begin{tabular}[t]{c}$n_{sims} = 5$\end{tabular}}}}%
    \put(0.37038587,0.45610229){\color[rgb]{0,0,0}\makebox(0,0)[t]{\lineheight{1.25}\smash{\begin{tabular}[t]{c}$n_{sims} = 20$\end{tabular}}}}%
    \put(0.59797688,0.45610229){\color[rgb]{0,0,0}\makebox(0,0)[t]{\lineheight{1.25}\smash{\begin{tabular}[t]{c}$n_{sims} = 50$\end{tabular}}}}%
    \put(0.82556795,0.45610229){\color[rgb]{0,0,0}\makebox(0,0)[t]{\lineheight{1.25}\smash{\begin{tabular}[t]{c}$n_{sims} = 100$\end{tabular}}}}%
    \put(0.95639911,0.44778132){\color[rgb]{0,0,0}\makebox(0,0)[t]{\lineheight{1.25}\smash{\begin{tabular}[t]{c}$N_c$\end{tabular}}}}%
  \end{picture}%
\endgroup%

%% file: chapters/related_work.tex
\section{Related Work}
\label{sec:related}

MCTS stands as a widely embraced algorithm, finding application across diverse domains \cite{SeglerPW18chemsynth,dieb_ju_shiomi_tsuda_2019}, prompting a multitude of extensions and enhancements \cite{browne_survey_2012,swiechowski_monte_2023}. Extending beyond general adaptations, the integration of domain-specific knowledge has gained prominence, exemplified by \cite{zheng_single-player_2021}. In the context of Sokoban \cite{crippa_analysis_2022}, employs cycle avoidance, tracing visited nodes and implementing recursive node elimination to enhance the algorithm. 

Our goal revolves around augmenting exploration by disregarding visited terminal nodes and completely explored subtrees. A highly related work regarding this is \citeauthor{moerland_monte_2018}~\shortcite{moerland_monte_2018,moerland_second_2020}. They reduce computational overhead for terminal nodes by modifying the UCT-term in a manner favoring actions with high uncertainty over those with high \qvalue{}s. This however, strongly influences the exploration-exploitation mechanism of MCTS and sacrificing some of its guarantees and characteristics. In contrast, we ensure a visit to terminal nodes only once, maintaining minimal alterations to the core MCTS algorithm, preserving its original principles and guarantees, and achieving substantial computational savings across a broader search space.

\citeauthor{windands_solver_2008}~\shortcite{windands_solver_2008} introduced the MCTS-Solver, enhancing MCTS in two-player zero-sum sudden-death games by introducing game-theoretic values ($\infty$ for wins and $-\infty$ for losses). While akin to our approach in implicitly pruning proven paths for losses, we distinguish ourselves by addressing a broader problem scope with simpler yet more effective changes. To the best of our knowledge, our extension marks a frontiering effort for a broader range of MDPs.

MCTS also gained popularity combined with neural networks in past years, notably exemplified by AlphaGo~\cite{silver2016mastering}, numerous advancements have been proposed \cite{swiechowski_monte_2023}. Correlating with works by \citeauthor{moerland_second_2020}~\shortcite{moerland_second_2020}, \citeauthor{lan_learning_2021}~\shortcite{lan_learning_2021} leverages uncertainty estimation to curtail the computational cost of MCTS. In our work we focused on the applicability of our changes and improvement in environments specifically developed to show problems with MCTS. While not evaluated in this work, we are already planning to test our changes with neural networks in such domains, e.g., chess or the game of Go.

%% file: chapters/limitations.tex
\section{Limitations}
\label{sec:limitations}
In our motivation, we have already stated that our improvements try to to maximize the use of MCTS-simulations for \textit{single-player deterministic discrete action space problems}. However, it does not solve exploration flaws other than the ones mentioned above.

Further, caused by the way transpositions are handled in our implementation, all rewards for non-terminal states must be positive, i.e.,
\begin{equation*}
    \mathcal{R}(s) \geq 0, \forall s: \lnot \, s.\mathit{terminal}.
\end{equation*}
This might be a restriction in theory, in practice however, a lot of tasks have a very sparse reward signal with only a reward signal in terminal states anyway, a punishment for taking more time to finish a task can easily be handled by the discount factor $\gamma$ instead of a steady negative reward signal.

%% file: chapters/conclusion.tex
\section{Conclusion}
\label{sec:discussion}

In conclusion, MCTS' strategic allocation of computational resources towards promising segments within a search tree makes it an appealing choice for exploration in vast search spaces. However, a notable drawback lies in its tendency to expend resources reevaluating previously explored regions. Addressing this limitation, our proposed methodology, labeled \ours{}, introduces a novel MCTS formulation. At the core of \ours{} is the innovative decoupling of value updates, visit count updates, and the selected path during tree search, allowing for the exclusion of already explored subtrees or leaves. It enables a significantly broader search with the same amount of computation, retaining the fundamental characteristics of MCTS. Therefore, it produces a much more accurate estimate of the true values, as reflected in the resulting performance which clearly outperforms the classical MCTS and other related work by a remarkable margin. As a result, \ours{} emerges as a promising enhancement to MCTS, offering improved efficiency and efficacy in navigating complex search spaces.

In future research, we hope to see groundbreaking results from applying this approach to solve practical problems. First, we want to combine our enhanced MCTS with AlphaZero-like approaches to improve decision making for single-player tasks but also in classical domains like chess endgame situations, before testing it in real-world applications like chemical syntheses or material design.

%% file: chapters/appendix.tex
\section{The Grammar for the Coverage of the Search Space Experiment}
\label{sec:grammar}
A context-free grammar \cite{parkes2008concise} is defined as a tuple $\mathfrak{G=(N,T,R,S)}$. 
$\mathfrak{N}$ is a set containing the non-terminal symbols.
For this symbols a production rule in the set $\mathfrak{R}$ exist of the form $B \rightarrow{} \beta{}$.
The production rule states that the non-terminal $B$ can be substituted by the terminal $\beta$, with $B \in \mathfrak{N}$ and $\beta \in ( \mathfrak{N} \cup \mathfrak{T})*$. 
$\mathfrak{T}$ is the set of terminal symbols, i.e., those symbols which cannot be expanded. 
Starting from a starting symbol $\mathfrak{S}$ an equation can be constructed by repeatedly applying the rules in $\mathfrak{R}$.
The following grammar is able to generate the equations from the well known Nguyen Symbolic Regression Benchmark Suite \cite{uy_semantically-based_2011} and is used in our experiment in \cref{sec:cov,fig:cov}:
\begin{align*} 
1.\ & \mathfrak{N} = \left\{ \mathit{Start}, \mathit{InnerFunction}, \mathit{Variable}, \mathit{Exponent}, \mathit{Sum} \right\} \\ 
2.\ & \mathfrak{T} \,= \left\{+, -, \cdot, \sin, \cos, \log, ^\wedge{}, x_0, x_1, 6, 5, 4, 3, 2, 1, 0.5\right\} \\
3.\ & \mathfrak{R} = \{\\
         &\quad \quad \quad \mathit{Start} \rightarrow{} 2\\
        &\quad \quad \quad \mathit{Start} \rightarrow{} 1\\
        &\quad \quad \quad \mathit{Start} \rightarrow{} 0.5\\
        &\quad \quad \quad \mathit{Start} \rightarrow{} + \mathit{Start} \, \mathit{Start}\\
        &\quad \quad \quad \mathit{Start} \rightarrow{} - \mathit{Start} \, \mathit{Start}\\
        &\quad \quad \quad \mathit{Start} \rightarrow{} \cdot \mathit{Start} \, \mathit{Start}\\
        &\quad \quad \quad \mathit{Start} \rightarrow{} \sin\mathit{ InnerFunction} \\
        &\quad \quad \quad \mathit{Start} \rightarrow{} \cos\mathit{ InnerFunction} \\
        &\quad \quad \quad \mathit{Start} \rightarrow{} \log \mathit{ InnerFunction} \\
        &\quad \quad \quad \mathit{Start} \rightarrow{} \mathit{Variable}\\
        &\quad \quad \quad \mathit{Start} \rightarrow{} ^\wedge{}\mathit{Exponent} \, \mathit{Variable} \\
        &\quad \quad \quad \mathit{Exponent} \rightarrow{} 6 \\
        &\quad \quad \quad \mathit{Exponent} \rightarrow{} 5 \\
        &\quad \quad \quad \mathit{Exponent} \rightarrow{} 4 \\
        &\quad \quad \quad \mathit{Exponent} \rightarrow{} 3 \\
        &\quad \quad \quad \mathit{Exponent} \rightarrow{} 2 \\
        &\quad \quad \quad \mathit{Exponent} \rightarrow{} 0.5 \\
        &\quad \quad \quad \mathit{Exponent} \rightarrow{} x_1 \\
        &\quad \quad \quad \mathit{InnerFunction} \rightarrow{} ^\wedge{}\mathit{Exponent} \, \mathit{Variable} \\
        &\quad \quad \quad \mathit{InnerFunction} \rightarrow{} x_0 \\
        &\quad \quad \quad \mathit{InnerFunction} \rightarrow{} x_1 \\
        &\quad \quad \quad \mathit{InnerFunction} \rightarrow{} + \mathit{Sum} \, \mathit{Sum} \\
        &\quad \quad \quad \mathit{Sum} \rightarrow{} ^\wedge{} \mathit{Exponent} \, \mathit{Variable} \\
        &\quad \quad \quad \mathit{Sum} \rightarrow{} 1 \\
        &\quad \quad \quad \mathit{Sum} \rightarrow{} x_0 \\
        &\quad \quad \quad \mathit{Sum} \rightarrow{} x_1 \\
        &\quad \quad \quad \mathit{Variable} \rightarrow{} x_0 \\
        &\quad \quad \quad \mathit{Variable} \rightarrow{} x_1\\
&\ \quad \quad\} \\
4.\ & \mathfrak{S} = \mathit{Start}.
\end{align*}

%% file: paper.bbl
\begin{thebibliography}{}

\bibitem[\protect\citeauthoryear{Brockman \bgroup \em et al.\egroup
  }{2016}]{brockman2016openai}
Greg Brockman, Vicki Cheung, Ludwig Pettersson, Jonas Schneider, John Schulman,
  Jie Tang, and Wojciech Zaremba.
\newblock Openai gym, 2016.

\bibitem[\protect\citeauthoryear{Browne \bgroup \em et al.\egroup
  }{2012}]{browne_survey_2012}
Cameron~B. Browne, Edward Powley, Daniel Whitehouse, Simon~M. Lucas, Peter~I.
  Cowling, Philipp Rohlfshagen, Stephen Tavener, Diego Perez, Spyridon
  Samothrakis, and Simon Colton.
\newblock A {Survey} of {Monte} {Carlo} {Tree} {Search} {Methods}.
\newblock {\em IEEE Transactions on Computational Intelligence and AI in
  Games}, 4(1):1--43, March 2012.

\bibitem[\protect\citeauthoryear{Coulom}{2007}]{coulom_efficient_2007}
Rémi Coulom.
\newblock Efficient {Selectivity} and {Backup} {Operators} in {Monte}-{Carlo}
  {Tree} {Search}.
\newblock In H.~Jaap van~den Herik, Paolo Ciancarini, and H.~H. L. M.~(Jeroen)
  Donkers, editors, {\em Computers and {Games}}, Lecture {Notes} in {Computer}
  {Science}, pages 72--83, Berlin, Heidelberg, 2007. Springer.

\bibitem[\protect\citeauthoryear{Crippa \bgroup \em et al.\egroup
  }{2022}]{crippa_analysis_2022}
Mattia Crippa, Pier~Luca Lanzi, and Fabio Marocchi.
\newblock An analysis of {Single}-{Player} {Monte} {Carlo} {Tree} {Search}
  performance in {Sokoban}.
\newblock {\em Expert Syst. Appl.}, 192(C), April 2022.

\bibitem[\protect\citeauthoryear{Devata \bgroup \em et al.\egroup
  }{2023}]{devata_deepspinn_2023}
Sriram Devata, Bhuvanesh Sridharan, Sarvesh Mehta, Yashaswi Pathak, Siddhartha
  Laghuvarapu, Girish Varma, and Deva Priyakumar.
\newblock {DeepSPInN} - multimodal {Deep} learning for molecular {Structure}
  {Prediction} from {Infrared} and {NMR} spectra.
\newblock preprint, Chemistry, November 2023.

\bibitem[\protect\citeauthoryear{Dieb \bgroup \em et al.\egroup
  }{2019}]{dieb_ju_shiomi_tsuda_2019}
Thaer~M. Dieb, Shenghong Ju, Junichiro Shiomi, and Koji Tsuda.
\newblock Monte carlo tree search for materials design and discovery.
\newblock {\em MRS Communications}, 9(2):532–536, 2019.

\bibitem[\protect\citeauthoryear{Kamienny \bgroup \em et al.\egroup
  }{2022}]{kamienny_end--end_2022}
Pierre-Alexandre Kamienny, St\'{e}phane d\textquotesingle Ascoli, Guillaume
  Lample, and Francois Charton.
\newblock End-to-end symbolic regression with transformers.
\newblock In S.~Koyejo, S.~Mohamed, A.~Agarwal, D.~Belgrave, K.~Cho, and A.~Oh,
  editors, {\em Advances in Neural Information Processing Systems}, volume~35,
  pages 10269--10281. Curran Associates, Inc., 2022.

\bibitem[\protect\citeauthoryear{Kocsis and
  Szepesvári}{2006}]{kocsis_bandit_2006}
Levente Kocsis and Csaba Szepesvári.
\newblock Bandit {Based} {Monte}-{Carlo} {Planning}.
\newblock In {\em European conference on machine learning}, volume 2006, pages
  282--293. Springer, September 2006.

\bibitem[\protect\citeauthoryear{Lan \bgroup \em et al.\egroup
  }{2021}]{lan_learning_2021}
Li-Cheng Lan, Ti-Rong Wu, I-Chen Wu, and Cho-Jui Hsieh.
\newblock Learning to {Stop}: {Dynamic} {Simulation} {Monte}-{Carlo} {Tree}
  {Search}.
\newblock {\em Proceedings of the AAAI Conference on Artificial Intelligence},
  35(1):259--267, May 2021.

\bibitem[\protect\citeauthoryear{Luo \bgroup \em et al.\egroup
  }{2022}]{luo_alphatruss_2022}
Ruifeng Luo, Yifan Wang, Weifang Xiao, and Xianzhong Zhao.
\newblock {AlphaTruss}: {Monte} {Carlo} {Tree} {Search} for {Optimal} {Truss}
  {Layout} {Design}.
\newblock {\em Buildings}, 12(5):641, May 2022.

\bibitem[\protect\citeauthoryear{Moerland \bgroup \em et al.\egroup
  }{2018}]{moerland_monte_2018}
Thomas~M. Moerland, Joost Broekens, Aske Plaat, and Catholijn~M. Jonker.
\newblock Monte {Carlo} {Tree} {Search} for {Asymmetric} {Trees}, May 2018.
\newblock arXiv:1805.09218 [cs, stat].

\bibitem[\protect\citeauthoryear{Moerland \bgroup \em et al.\egroup
  }{2020}]{moerland_second_2020}
Thomas~M. Moerland, Joost Broekens, Aske Plaat, and Catholijn~M. Jonker.
\newblock The {Second} {Type} of {Uncertainty} in {Monte} {Carlo} {Tree}
  {Search}, May 2020.
\newblock arXiv:2005.09645 [cs].

\bibitem[\protect\citeauthoryear{Nunes \bgroup \em et al.\egroup
  }{2018}]{nunes_monte_2018}
Cecilia Nunes, Mathieu De~Craene, Helene Langet, Oscar Camara, and Anders
  Jonsson.
\newblock A {Monte} {Carlo} {Tree} {Search} {Approach} to {Learning} {Decision}
  {Trees}.
\newblock In {\em 2018 17th {IEEE} {International} {Conference} on {Machine}
  {Learning} and {Applications} ({ICMLA})}, pages 429--435, Orlando, FL,
  December 2018. IEEE.

\bibitem[\protect\citeauthoryear{Parkes}{2008}]{parkes2008concise}
Alan~P Parkes.
\newblock {\em A concise introduction to languages and machines}.
\newblock Springer Science \& Business Media, 2008.

\bibitem[\protect\citeauthoryear{Petersen \bgroup \em et al.\egroup
  }{2021}]{petersen_deep_2021}
Brenden~K. Petersen, Mikel Landajuela, T.~Nathan Mundhenk, Claudio~P. Santiago,
  Soo~K. Kim, and Joanne~T. Kim.
\newblock Deep symbolic regression: {Recovering} mathematical expressions from
  data via risk-seeking policy gradients, April 2021.
\newblock arXiv:1912.04871 [cs, stat].

\bibitem[\protect\citeauthoryear{Segler \bgroup \em et al.\egroup
  }{2018}]{SeglerPW18chemsynth}
Marwin H.~S. Segler, Mike Preuss, and Mark~P. Waller.
\newblock Planning chemical syntheses with deep neural networks and symbolic
  {AI}.
\newblock {\em Nat.}, 555(7698):604--610, 2018.

\bibitem[\protect\citeauthoryear{Silver \bgroup \em et al.\egroup
  }{2016}]{silver2016mastering}
David Silver, Aja Huang, Chris~J Maddison, Arthur Guez, Laurent Sifre, George
  Van Den~Driessche, Julian Schrittwieser, Ioannis Antonoglou, Veda
  Panneershelvam, Marc Lanctot, et~al.
\newblock Mastering the game of go with deep neural networks and tree search.
\newblock {\em nature}, 529(7587):484--489, 2016.

\bibitem[\protect\citeauthoryear{Silver \bgroup \em et al.\egroup
  }{2018}]{silver_general_2018}
David Silver, Thomas Hubert, Julian Schrittwieser, Ioannis Antonoglou, Matthew
  Lai, Arthur Guez, Marc Lanctot, Laurent Sifre, Dharshan Kumaran, Thore
  Graepel, Timothy Lillicrap, Karen Simonyan, and Demis Hassabis.
\newblock A general reinforcement learning algorithm that masters chess, shogi,
  and {Go} through self-play.
\newblock {\em Science}, 362(6419):1140--1144, December 2018.
\newblock Publisher: American Association for the Advancement of Science.

\bibitem[\protect\citeauthoryear{Sridharan \bgroup \em et al.\egroup
  }{2021}]{sridharan_spectra_2021}
Bhuvanesh Sridharan, Sarvesh Mehta, Yashaswi Pathak, and U.~Deva Priyakumar.
\newblock Spectra to {Structure}: {Deep} {Reinforcement} {Learning} for
  {Molecular} {Inverse} {Problem}.
\newblock preprint, Chemistry, December 2021.

\bibitem[\protect\citeauthoryear{Uy \bgroup \em et al.\egroup
  }{2011}]{uy_semantically-based_2011}
Nguyen~Quang Uy, Nguyen~Xuan Hoai, Michael O’Neill, R.~I. McKay, and Edgar
  Galván-López.
\newblock Semantically-based crossover in genetic programming: application to
  real-valued symbolic regression.
\newblock {\em Genet Program Evolvable Mach}, 12(2):91--119, June 2011.

\bibitem[\protect\citeauthoryear{Winands \bgroup \em et al.\egroup
  }{2008}]{windands_solver_2008}
Mark H.~M. Winands, Yngvi Bj{\"o}rnsson, and Jahn-Takeshi Saito.
\newblock Monte-carlo tree search solver.
\newblock In H.~Jaap van~den Herik, Xinhe Xu, Zongmin Ma, and Mark H.~M.
  Winands, editors, {\em Computers and Games}, pages 25--36, Berlin,
  Heidelberg, 2008. Springer Berlin Heidelberg.

\bibitem[\protect\citeauthoryear{Zheng \bgroup \em et al.\egroup
  }{2021}]{zheng_single-player_2021}
Guangcong Zheng, Cong Wang, Weijie Shao, Ying Yuan, Zejie Tian, Sancheng Peng,
  Ali~Kashif Bashir, and Shahid Mumtaz.
\newblock A single-player {Monte} {Carlo} tree search method combined with node
  importance for virtual network embedding.
\newblock {\em Annals of Telecommunications}, 76(5-6):297--312, June 2021.

\bibitem[\protect\citeauthoryear{Świechowski \bgroup \em et al.\egroup
  }{2023}]{swiechowski_monte_2023}
Maciej Świechowski, Konrad Godlewski, Bartosz Sawicki, and Jacek Mańdziuk.
\newblock Monte {Carlo} {Tree} {Search}: a review of recent modifications and
  applications.
\newblock {\em Artificial Intelligence Review}, 56(3):2497--2562, March 2023.

\end{thebibliography}
